%% file: main.tex
\documentclass{article}
\usepackage{deeprobot_paper,times}
\usepackage[T1]{fontenc}
\usepackage[utf8]{inputenc}
\usepackage{hyperref}
\usepackage{amsmath,amssymb}
\usepackage{booktabs,array,multirow,colortbl,float}
\usepackage{microtype}
\usepackage{caption}
\newcommand{\method}{DR-VLA}

\title{DR-VLA: Learning Human-Like Quadruped\\Navigation from Egocentric Web Videos via\\Domain-Invariant Scene Representations}
\author{Jun Tian\textsuperscript{1,2}\quad Yingjie Niu\textsuperscript{1,*,\textdagger}\quad Ruo Chen\textsuperscript{3}\quad Hongyi Lin\textsuperscript{4,5}\quad Chao Li\textsuperscript{1,*}}
\affiliation{%
\textsuperscript{1}Deep Robotics\quad
\textsuperscript{2}University of Illinois at Urbana-Champaign\\
\textsuperscript{3}Zhejiang University\quad
\textsuperscript{4}Tsinghua University\\
\textsuperscript{5}Massachusetts Institute of Technology}
\authornote{\textsuperscript{*}Corresponding authors.\quad\textsuperscript{\textdagger}Project Leader.}
\paperdate{}
\projecturl{}
\hypersetup{%
  pdftitle={DR-VLA: Learning Human-Like Quadruped Navigation from Egocentric Web Videos via Domain-Invariant Scene Representations},
  pdfauthor={Jun Tian, Yingjie Niu, Ruo Chen, Hongyi Lin, Chao Li},
  pdfsubject={Quadruped navigation and human-to-robot transfer},
  pdfkeywords={DR-VLA, EgoLynx, quadruped navigation, world action model, DISR},
  bookmarksnumbered=true}

\begin{document}
\makebrandheader
\begin{deeprobotfrontmatter}
\maketitle
\begin{abstract}
Quadruped robots performing outdoor deliveries must interpret complex scenes, anticipate changes, and navigate in socially compliant ways. 
Robot demonstrations are costly to collect at the scale needed for generalization. Egocentric web videos offer abundant human navigation experience, but differences in viewpoint and motion make this experience difficult to transfer to robots. 
We introduce \method{}, a Mixture-of-Transformers framework that transfers human navigation experience through a domain-invariant scene representation (DISR). 
The framework combines vision-language understanding and latent world modeling with a flow-matching action expert.
We introduce EgoLynx, comprising 10,000 hours of curated egocentric web video (EgoLynx-Web) and 2,000 clips of teleoperated robot navigation collected during real-world deliveries (EgoLynx-Robot). 
We pretrain the understanding and world experts on EgoLynx-Web without action labels, then adapt the model to robot observations and train the action expert on EgoLynx-Robot to predict executable trajectories. 
We also propose the Quadruped Navigation Score (QNS), a pseudo-closed-loop metric combining safety, instruction compliance, comfort, and progress. 
On the held-out EgoLynx-Robot test split, \method{} improves QNS by approximately 5\% and reduces open-loop final heading error (FHE) by 14.6\%, each relative to the best competing result for that metric. 
Ablations show that web-video pretraining improves generalization and that the model’s key components contribute to performance. As a future-prediction target, DISR outperforms five generic visual representations and yields performance close to task-specific 3D perception without dedicated geometric labels or point clouds for target construction.
Trained separately on NAVSIM v1 without web pretraining, \method{} achieves 92.1 PDMS, suggesting that the framework also applies to autonomous driving.

\end{abstract}
\end{deeprobotfrontmatter}

\begin{figure}[htbp]
\centering
\includegraphics[width=0.98\textwidth]{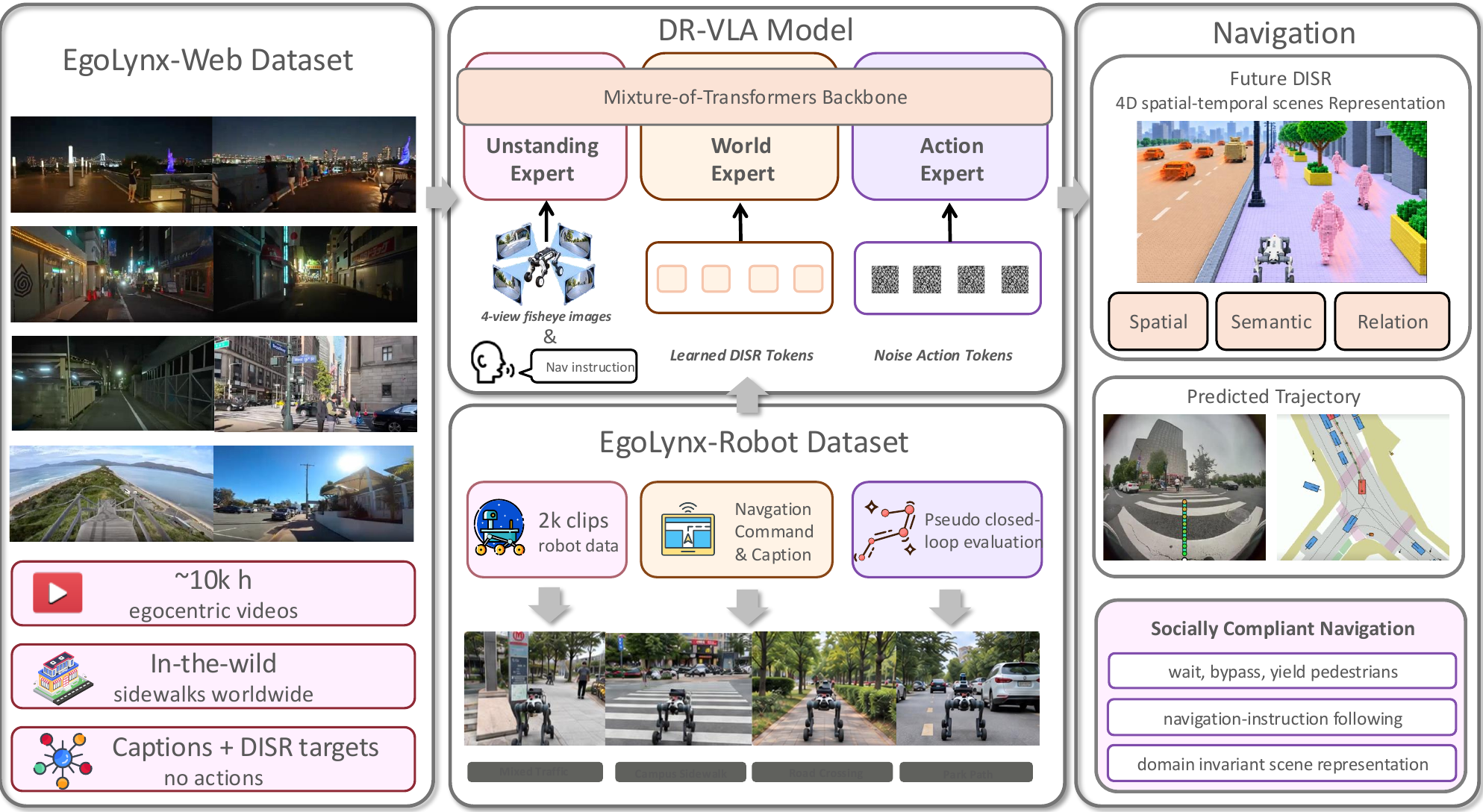}
\caption{\textbf{Overview.} \method{} learns to understand and anticipate scenes from egocentric web videos (left) and to act from teleoperated quadruped clips (center bottom). Three experts (center top) map four fisheye views and a navigation instruction to a trajectory and to future domain-invariant scene representations (DISR), which encode semantic, spatial, and relational structure (right).}
\label{fig:teaser}
\end{figure}

\section{Introduction}
\label{sec:intro}
Last-mile delivery in shared outdoor spaces, from city sidewalks to village paths, is a promising application for quadruped robots \citep{alverhed2024lastmile,lee2024wheeledlegged}. 
Navigating these spaces requires more than reaching a goal without collision. 
A robot must understand complex scenes, anticipate how the environment will change, and navigate in socially compliant ways. 
The prevailing approach learns these capabilities end-to-end from robot demonstrations via imitation learning \citep{gnm,vint,nomad,omnivla}. 
However, collecting demonstrations requires skilled teleoperation and careful safety oversight \citep{citywalker}. 
As a result, existing quadruped data are limited in both scene coverage and diversity of social interactions: even the largest public real-world datasets contain only 5--15 hours of legged-robot data each, typically from a single city or region \citep{scand,citywalker,grandtour}. 
With such data alone, models struggle to cover the long-tail situations that these three capabilities must handle.

In contrast, egocentric videos on the Internet offer another source of data. 
They are easy to obtain and span cities and weather conditions worldwide. 
More importantly, they show how people move through varied environments, respond to other agents, and choose paths around obstacles, which is exactly what robot data lack the most. 
However, transferring this experience to robots presents two challenges.
The first is an observation gap: head-mounted human videos and a robot's low-mounted, multi-camera fisheye observations differ in viewpoint, imaging characteristics, and camera motion.
The second is an action gap: Internet videos do not provide robot action labels, and human motion cannot be directly executed by a quadruped. 
Prior work estimates pseudo-actions from camera motion or reconstructed scene geometry \citep{citywalker,navila,kuno,vega}. 
These methods help address missing action labels, but the resulting supervision may retain human viewpoint and motion biases. 
Other approaches learn from unlabeled videos by predicting future visual content \citep{nwm,aether}. 
These methods learn scene dynamics, but they entangle embodiment-specific factors such as appearance and viewpoint with scene evolution, which hinders transfer across embodiments. 
Some works transfer at an embodiment-agnostic abstract level \citep{egonav}, but they rely on depth and pose captured with dedicated wearable devices and thus cannot scale to Internet video. 
Existing methods therefore face a dilemma: they either scale but transfer at an embodiment-specific level, or use abstract representations but fail to scale. 
These limitations motivate a central question: can we learn transferable scene understanding and future prediction directly from raw Internet videos, at an abstract level shared by humans and robots?

We address this question with a domain-invariant scene representation (DISR) designed to retain information useful for navigation across human and robot observations. 
We observe that scene understanding for outdoor navigation comes down to semantics, space, and relations: what is in the scene, where it is, and how entities relate to one another. 
We obtain semantic information from segmentation and spatial structure from depth, then fuse them with attention to represent relations among entities and to the surrounding geometry. 
A pretrained scene autoencoder compresses this information into a compact latent. 
Segmentation and depth can be estimated from RGB video using vision foundation models \citep{sam3,depthanything3}, allowing DISR targets to be generated without dedicated depth sensors or manual action annotation. 
By emphasizing scene structure over texture and imaging style, DISR reduces appearance-related differences between human and robot observations. Adaptation on robot data addresses the remaining viewpoint and embodiment differences.
We model scene dynamics by predicting future DISR latents rather than reconstructing future pixels. 
Future observations in human videos reflect the consequences of navigation decisions, such as moving around a crowd or stopping to yield. 
Predicting their scene-level evolution thus provides a way to learn human navigation priors without explicit action labels.

Building on this insight, we propose \method{}, a Mixture-of-Transformers \citep{mot} framework whose three experts have separate parameters and exchange information via shared self-attention. 
This tightly couples understanding, prediction, and decision-making while avoiding interference among training objectives. 
The understanding expert is a vision-language model that takes images and navigation instructions as input and reasons about the scene under chain-of-thought-style caption supervision \citep{cot,ecot}. 
The world expert predicts future DISR latents. Conditioned on the other experts, the action expert generates robot trajectories through flow matching \citep{flowmatching}.
To support training across human and robot observations, we curate EgoLynx from complementary data sources. 
EgoLynx-Web contains 10,000 hours of egocentric Internet video selected to cover diverse environments and human navigation behaviors. 
EgoLynx-Robot contains 2,000 training clips (approximately 17 hours) and a held-out test split, both selected from a larger collection of human-teleoperated wheeled-legged robot navigation in real-world delivery settings. 
We first pretrain the understanding and world experts on EgoLynx-Web without action labels, then adapt the model to robot observations, and supervise the action expert with EgoLynx-Robot.

We evaluate \method{} on real-world quadruped navigation and on the NAVSIM autonomous-driving benchmark \citep{navsim}. 
We also introduce the Quadruped Navigation Score (QNS) to evaluate navigation quality beyond agreement with a recorded trajectory. 
We report it alongside open-loop average and final displacement errors (ADE and FDE) on the held-out EgoLynx-Robot test split. Against eight navigation and vision-language-action baselines, \method{} improves QNS by approximately 5\% and reduces FHE by 14.6\%, each relative to the strongest competing result for that metric.
Ablations show that web-video pretraining improves performance on held-out robot data and that DISR is a more effective future-prediction target than five generic visual representations. A further comparison with task-specific 3D perception shows that DISR yields close target-only performance without Box/Line annotations or point clouds for target construction. Using perception as an online policy input improves QNS at additional computation cost (Sec.~\ref{sec:exp:perception}).
Trained independently on NAVSIM v1 without EgoLynx-Web pretraining, \method{} achieves 92.1 PDMS, and removing the world expert reduces the score to 89.5.

In summary, our contributions are fivefold:
(i) We propose a domain-invariant scene representation that abstracts human and robot observations into shared semantic, spatial, and relational structure, allowing human navigation priors such as social compliance to be distilled from large-scale egocentric video and transferred across embodiments.
(ii) We present \method{}, a Mixture-of-Transformers latent world action model whose understanding, world, and action experts decouple supervision across data sources and extend readily to diverse navigation settings.
(iii) We curate EgoLynx, a dataset that pairs 10k hours of egocentric web videos with 2,000 teleoperated quadruped clips under a shared annotation format of instructions, captions, and DISR targets, enabling action-free pretraining followed by data-efficient adaptation.
(iv) We introduce QNS, a pseudo-closed-loop quadruped navigation metric that evaluates safety, instruction compliance, comfort, and progress.
(v) \method{} outperforms state-of-the-art methods by approximately 5\% in QNS on real-world quadruped navigation and achieves 92.1 PDMS on NAVSIM, corroborating the generality of the representation across embodiments.

\section{Related Work}
\label{sec:related}

\textbf{VLA and WAM.} Vision-language-action (VLA) models adapt pretrained vision-language models to robot control \citep{rt2,openvla,pi0,pi05,grootn1}, and recent variants adopt Mixture-of-Transformers (MoT) designs whose function-specific experts share attention \citep{mot,pi0,f1vla,motus}. World action models (WAMs) \citep{dreamzero} additionally predict future observations: early ones forecast pixels or video latents \citep{unipi,gr1,vpp,uwm,uva,cotvla}, spending capacity on visual detail irrelevant to control, whereas recent ones predict latent embeddings of future observations as training-only targets \citep{flare,dreamvla,vlajepa,lda1b,beingh07}. \method{} follows this latent route but replaces generic visual embeddings, which retain the recording embodiment's appearance, with a structured scene representation that removes appearance by construction, and partitions its experts by supervision source.

\textbf{Visual and Socially Compliant Navigation.} Visual navigation foundation models transfer across platforms \citep{gnm,vint,nomad} and have been extended with web video \citep{lelan,citywalker} and VLA formulations \citep{navid,uninavid,omnivla}, including for legged robots \citep{navila,quarvla,germ,quartonline,more}. Social navigation further requires following implicit human conventions \citep{socialnavsurvey}, which prior work encodes through hand-crafted models or reinforcement learning in simulated crowds \citep{socialforce,orca,cadrl,sarl}, small-scale demonstrations \citep{scand,musohu}, or VLM-based reasoning \citep{vlmsocialnav,socialllava}, yet current VLMs still trail human consensus in social scene understanding \citep{socialnavsub}. \method{} instead learns such conventions from large-scale human video and explicitly bridges the human-to-quadruped gap that trajectory-level imitation leaves open.

\textbf{Learning from Egocentric Human Videos.} Egocentric datasets \citep{ego4d,egodex,sekai} have made human video a practical source of supervision. Beyond visual pretraining \citep{r3m,vip}, one line of work recovers pseudo-actions through pose retargeting \citep{egomimic,egovla,beingh0} or visual odometry \citep{citywalker,navila,kuno}. Other methods learn latent actions from inter-frame change \citep{lapa,univla,moto,villax}, which tends to entangle scene dynamics with camera motion unless explicitly separated. Another approach jointly aligns human and robot latents with actions \citep{egobridge}, requiring action-labeled human data. Most closely related, EgoWAM \citep{egowam} shows in a controlled manipulation study that the world-prediction target governs human-to-robot transfer and that pixel targets transfer weakly, yet finds no off-the-shelf target uniformly superior. \method{} builds on this premise with a dedicated target combining semantic, geometric, and relational structure, inspired by mid-level representations that aid transfer in navigation and driving \citep{mueller2018,midlevel,zhou2019}. Unlike these approaches, our standard model uses the representation only as a prediction target. We focus on navigation, where ego-motion differs on both sides of the gap.

\section{Task Formulation}
\label{sec:task}

\textbf{Navigation policy.}
We cast outdoor navigation as instruction-conditioned trajectory prediction. At time $t$, the agent receives an observation $\mathbf{o}_t = \{I^{v}_{t-T+1}, \dots, I^{v}_{t}\}_{v=1}^{V}$ comprising the last $T$ frames from $V$ camera views, a natural-language navigation instruction $\ell$, and its ego state $s_t$. The policy outputs an action chunk of $N_a$ planar body-frame poses spaced $\delta$ apart,
\begin{equation}
\mathbf{a}_t \sim \pi(\cdot \mid \mathbf{o}_t, \ell, s_t), \qquad
\mathbf{a}_t = \{(x_j, y_j, \theta_j)\}_{j=1}^{N_a}.
\label{eq:policy}
\end{equation}
On the quadruped, $V {=} 4$ surround-view fisheye cameras observe the scene, $\ell$ specifies the route (e.g., \emph{follow the path}, \emph{branch left}), and $\mathbf{a}_t$ spans 1~s with $N_a{=}10$ and $\delta{=}0.1$~s. In driving, $\ell$ is one of four high-level commands and $\mathbf{a}_t$ spans 4~s with $N_a{=}8$ and $\delta{=}0.5$~s. Only $V$, $N_a$, and $\delta$ depend on the embodiment. Importantly, $\ell$ prescribes \emph{where} to go but not \emph{how} to behave along the way: when to stop, yield, or detour around other agents is left unspecified and must be inferred from the scene.

\textbf{A structural view of the embodiment gap.}
Humans, quadrupeds, and vehicles traverse similar scenes, yet they perceive and act in them differently. We model an observation from embodiment $e$ as a rendering of an underlying scene state,
\begin{equation}
\mathbf{o}_t = \Gamma_e\big(\boldsymbol{\zeta}_t;\, \boldsymbol{\eta}_e\big),
\label{eq:render}
\end{equation}
where $\boldsymbol{\zeta}_t$ captures the scene structure relative to the observer, namely what is present, where it lies, and how entities relate, and $\boldsymbol{\eta}_e$ collects embodiment-specific factors such as camera height, appearance, and low-level motion characteristics. Under this view, a human egocentric recording is simply another embodiment with $V{=}1$. The gap between embodiments then arises at two levels: $\Gamma_e$ renders the same scene differently, and the action spaces differ, so human motion does not define an action $\mathbf{a}_t$ in the robot's space. We assume, however, that the evolution of scene structure, $p(\boldsymbol{\zeta}_{t+\Delta:t+K\Delta} \mid \boldsymbol{\zeta}_{\le t}, \ell)$, is largely shared across embodiments. Since $\boldsymbol{\zeta}$ is expressed relative to the observer, its evolution also reflects where the observer chooses to go, and hence encodes human navigation priors such as yielding to pedestrians or steering around crowds.

\textbf{Policy decomposition.}
This view suggests factorizing the policy into a scene-evolution predictor $\Psi$ shared across embodiments and an embodiment-specific action generator $\pi_e$,
\begin{equation}
\hat{\boldsymbol{\zeta}}_{t+\Delta:t+K\Delta} = \Psi(\mathbf{o}_t, \ell), \qquad
\mathbf{a}_t \sim \pi_e\big(\cdot \mid \mathbf{o}_t, \ell, s_t, \hat{\boldsymbol{\zeta}}_{t+\Delta:t+K\Delta}\big).
\label{eq:decomp}
\end{equation}
Because $\Psi$ depends only on scene structure, it can be learned from observations alone, including those of embodiments whose actions are unavailable or incompatible with the robot. In contrast, $\pi_e$ grounds the anticipated scene evolution in the robot's own kinematics and therefore requires action supervision from the target embodiment.

\section{Methodology}
\label{sec:method}

\begin{figure}[t]
\centering
\includegraphics[width=0.98\textwidth]{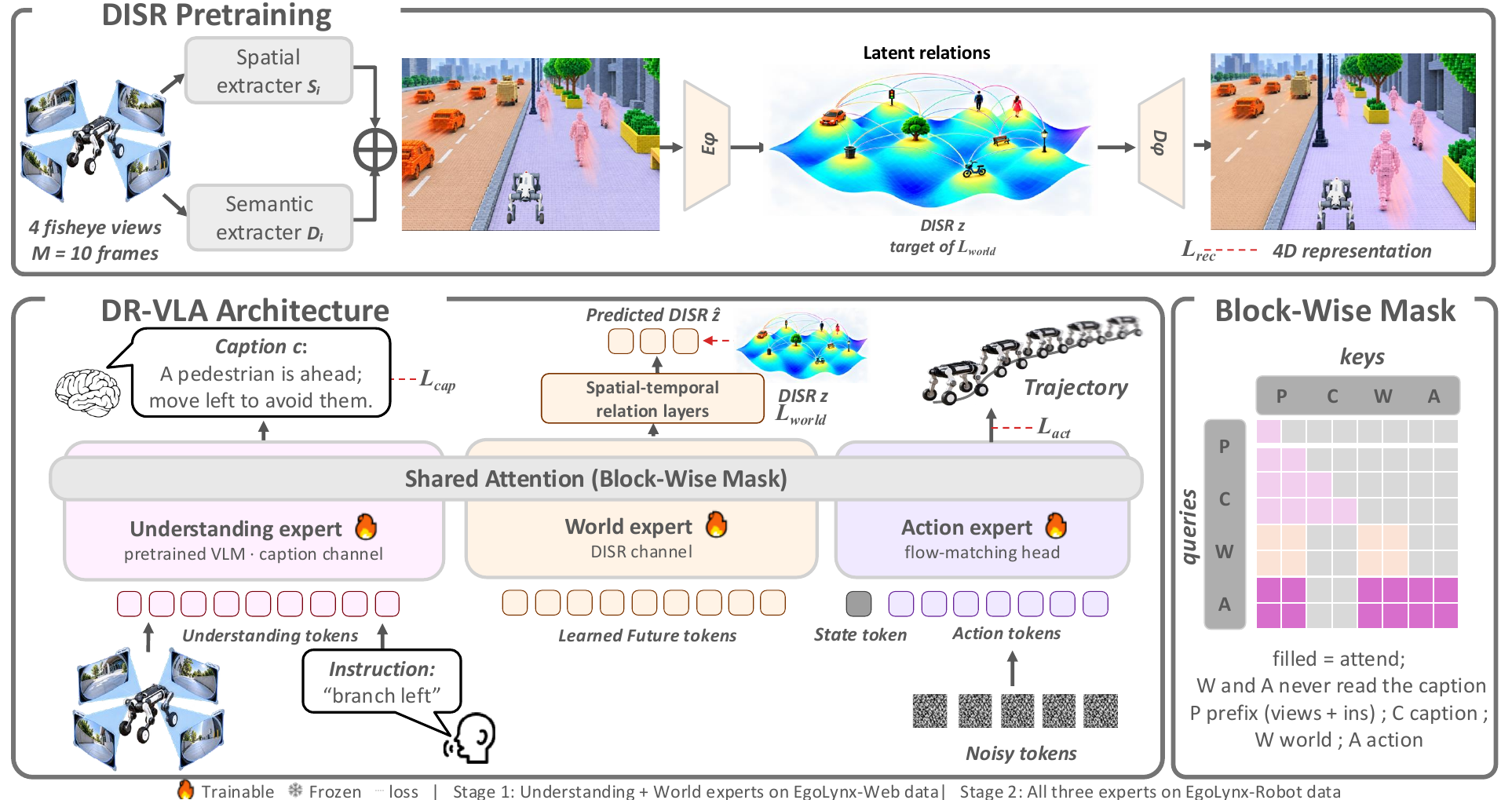}
\caption{\textbf{\method{} and the DISR Pretraining.} \emph{Top:} Combined Spatial and Semantic information are encoded by $E_\phi$ into the DISR latent $\mathbf{z}$ and reconstructed by $D_\phi$. \emph{Bottom left:} the understanding, world, and action experts share attention and are trained with $\mathcal{L}_{\text{cap}}$, $\mathcal{L}_{\text{world}}$ (targets from the frozen $E_\phi$), and $\mathcal{L}_{\text{act}}$. \emph{Bottom right:} block-wise attention mask.}
\label{fig:arch}
\end{figure}

\subsection{Domain-Invariant Scene Representation}
\label{sec:disr}

We define the domain-invariant scene representation (DISR) as a compact spatio-temporal latent $\mathbf{z}$ that serves as a tractable approximation of the scene structure $\boldsymbol{\zeta}$. DISR characterizes a scene along three axes: \emph{semantic}, \emph{spatial}, and \emph{relational}. Vision foundation models extract the first two explicitly from raw frames, while an encoder learns the third across frames (Fig.~\ref{fig:arch}, top).

\paragraph{Disentangling structure from appearance.}
For each frame $I_i$ of each view, two frozen teachers provide structural cues. The semantic cue $S_i$ is produced by the concept segmentation model SAM~3 \citep{sam3} over a fixed category vocabulary $\mathcal{C}_{\text{sem}}$ shared across all data (e.g., pedestrian, vehicle), and is encoded as $|\mathcal{C}_{\text{sem}}|$ channels of category masks. The spatial cue $D_i$ is the relative depth estimated by the relative-depth model of Depth Anything~3 \citep{depthanything3}. Discarding absolute scale also mitigates the scale discrepancy induced by different camera heights. Together, the image-plane layout of the masks and the depth locate each entity in direction and distance relative to the observer. Both teachers operate directly on raw frames, including unrectified fisheye images. Because the encoder receives only $(S_i, D_i)$ and never RGB, appearance factors such as texture, lighting, and imaging style are excluded by construction.

\paragraph{Relational encoding in a shared latent space.}
The DISR encoder $E_\phi$ and a lightweight decoder $D_\phi$ form an autoencoder. For $M{=}10$ consecutive frames, $S_i$ and $D_i$ are concatenated along the channel dimension, tokenized, and processed by $L_{\text{rel}}$ layers of causal spatio-temporal self-attention over the tokens of all frames,
\begin{equation}
\{\mathbf{z}_1, \dots, \mathbf{z}_M\} = E_\phi\big(\{(S_i, D_i)\}_{i=1}^{M}\big), \qquad \mathbf{z}_i \in \mathbb{R}^{N_z \times d_z}.
\label{eq:disr}
\end{equation}
These attention layers realize the relational axis, capturing entity correspondence across frames, relative depth ordering, and co-occurrence among entities. The decoder reconstructs $S_i$ and $D_i$ from each $\mathbf{z}_i$ under a reconstruction loss plus a regularizer $\mathcal{R}(\mathbf{z}_i)$. Compared with discrete label maps, the continuous latent is more compact and better suited as a regression target.

We train $E_\phi$ on a mixture of human videos and robot data, then freeze it and apply the same weights to clips from every embodiment, mapping all observations into a common target space. In the standard \method{} configuration, DISR serves only as the training target of the world expert (Sec.~\ref{sec:model}), so neither the teachers nor the DISR encoder is needed at inference. We separately evaluate DISR as a policy input in Sec.~\ref{sec:exp:perception}.

\subsection{Mixture-of-Transformers Architecture}
\label{sec:model}

We instantiate the decomposition in Eq.~(\ref{eq:decomp}) as \method{} (Fig.~\ref{fig:arch}). The understanding and world experts jointly realize $\Psi$: the former grounds the instruction and multi-view observations in scene semantics, and the latter predicts the DISR of $K$ future frames. The action expert realizes $\pi_e$, translating the anticipated scene evolution into trajectories for the robot's own body. Following the Mixture-of-Transformers design \citep{pi0,mot,bagel}, each expert owns its attention projections. Image tokens of view $v$ receive a learnable view embedding $e_v$ added to their positional encoding, which allows the model to distinguish the four fisheye views without rectification.

\textbf{Understanding expert: grounding instructions in the scene.}
The understanding expert $f_u$ is a pretrained vision-language model, Qwen3-VL-2B \citep{qwen3vl}. Images from all views are encoded by its ViT and the instruction $\ell$ by its text tokenizer, together forming the prefix tokens. During training, $f_u$ generates the clip caption $c$ by next-token prediction,
\begin{equation}
\mathcal{L}_{\text{cap}} = -\sum_{n} \log p_{f_u}\big(c_n \mid c_{<n}, \mathbf{o}_t, \ell\big).
\label{eq:cap}
\end{equation}
The caption describes the scene layout, surrounding agents, potential hazards, and ego-motion, steering the hidden states $\mathbf{h}^u$ toward navigation-relevant content. It serves only as auxiliary supervision and is not decoded at inference.

\textbf{World expert: anticipating scene evolution.}
The world expert $f_w$ operates on learnable world tokens $\mathbf{w} = \{w_1, \dots, w_K\}$, where $w_k$ holds $N_w$ query tokens per view, augmented with a temporal embedding for time $t+k\Delta$ and the view embedding $e_v$. The world tokens first read the prefix through shared attention and then pass through $L_w$ spatio-temporal relation layers spanning all world tokens. These layers relate the predicted frames and views, extending the relational axis of DISR into the future. With $N_w = N_z$, a token-wise linear head $g_w$ yields $\hat{\mathbf{z}}_{t+k\Delta} = g_w(\mathbf{h}^w_k) \in \mathbb{R}^{N_z \times d_z}$. The target $\mathbf{z}_{t+k\Delta}$ is computed by the frozen $E_\phi$ on the causal clip ending at frame $t+k\Delta$ (Sec.~\ref{sec:disr}). Abbreviating $\hat{\mathbf{z}}_{t+k\Delta}$ and $\mathbf{z}_{t+k\Delta}$ as $\hat{\mathbf{z}}_k$ and $\mathbf{z}_k$, the world loss is
\begin{equation}
\mathcal{L}_{\text{world}} = \frac{1}{K}\sum_{k=1}^{K}\Big[d_{\text{SL1}}(\hat{\mathbf{z}}_k, \mathbf{z}_k) + 1 - \cos(\hat{\mathbf{z}}_k, \mathbf{z}_k)\Big],
\label{eq:world}
\end{equation}
where $d_{\text{SL1}}$ denotes the smooth-L1 distance and $\cos$ the mean token-wise cosine similarity.

\textbf{Action expert: embodiment-specific trajectory generation.}
The action expert $f_a$ generates the action chunk $\mathbf{a}_t$ via conditional flow matching \citep{flowmatching,pi0}. Each of the $N_a$ waypoints is represented by one action token, and the ego state $s_t$ occupies one additional token. Given $\tau \sim \mathcal{U}[0,1]$, $\boldsymbol{\epsilon} \sim \mathcal{N}(\mathbf{0}, \mathbf{I})$, and $\mathbf{a}_t^{\tau} = \tau\,\mathbf{a}_t + (1-\tau)\,\boldsymbol{\epsilon}$, $f_a$ predicts the velocity field under
\begin{equation}
\mathcal{L}_{\text{act}} = \mathbb{E}_{\tau, \boldsymbol{\epsilon}}\Big\| v_{f_a}\big(\mathbf{a}_t^{\tau}, \tau, s_t;\, \mathbf{h}^u, \mathbf{h}^w\big) - (\mathbf{a}_t - \boldsymbol{\epsilon}) \Big\|_2^2.
\label{eq:act}
\end{equation}
The action expert attends to $\mathbf{h}^u$ and $\mathbf{h}^w$ through shared attention. Since $\hat{\mathbf{z}}$ is read out from $\mathbf{h}^w$ by $g_w$, conditioning on $\mathbf{h}^w$ realizes the dependence of $\pi_e$ on the anticipated scene evolution in Eq.~(\ref{eq:decomp}). At inference, $\mathbf{a}_t$ is integrated from noise in 4 Euler steps.

\textbf{Structured information flow.}
A block-wise attention mask \citep{pi0} governs how information flows among tokens (Fig.~\ref{fig:arch}, bottom right). World and action tokens do not attend to the caption, so their inputs are identical at training and inference. Moreover, since action tokens are not visible to the prefix or world tokens, $\mathbf{h}^u$ and $\mathbf{h}^w$ are computed once at inference and reused across integration steps.

\subsection{EgoLynx and Two-Stage Training}
\label{sec:model:training}

To instantiate the decomposition in Eq.~(\ref{eq:decomp}), we curate EgoLynx, a dataset that pairs action-free human videos with action-labeled robot demonstrations under a shared annotation format: every clip carries an instruction $\ell$, a caption $c$, and DISR targets $\mathbf{z}$, while robot clips additionally carry actions $\mathbf{a}$. Its two parts, EgoLynx-Web and EgoLynx-Robot, drive two training stages that learn $\Psi$ and $\pi_e$, respectively, with the DISR encoder $E_\phi$ kept frozen throughout.

\textbf{Stage 1: learning scene evolution from EgoLynx-Web.}
EgoLynx-Web is collected from YouTube and covers walking, cycling, and other forms of egocentric human navigation across urban and rural areas, different times of day, and diverse weather conditions. After multi-stage cleaning, it retains about 10k hours, roughly twice the size of Sekai and three times that of Ego4D \citep{sekai,ego4d}. For each clip, Gemini~3.1 Flash-Lite \citep{gemini} observes the full clip, including its subsequent motion, and generates the instruction $\ell$ and the caption $c$. This stage trains only $f_u$ and $f_w$ with $\mathcal{L}_1 = \mathcal{L}_{\text{cap}} + \lambda_w \mathcal{L}_{\text{world}}$. We deliberately avoid estimating pseudo-trajectories from the videos: human motion does not correspond to the robot's action space, and camera motion estimated on monocular, dynamic footage is noisy \citep{megasam}.

\textbf{Stage 2: grounding in the target embodiment with EgoLynx-Robot.}
EgoLynx-Robot comprises 2,000 training teleoperated clips (approximately 17 hours) selected from a larger collection of human-teleoperated wheeled-legged robot navigation in real-world delivery settings. Its captions are generated with the same prompt as in Stage~1, grounded in the event annotations. This stage initializes the action expert and the view embeddings of the four fisheye cameras, and fine-tunes all parameters with $\mathcal{L}_2 = \mathcal{L}_1 + \lambda_a \mathcal{L}_{\text{act}}$. Retaining $\mathcal{L}_{\text{cap}}$ and $\mathcal{L}_{\text{world}}$ guards against the erosion of the understanding and prediction capabilities acquired in Stage~1 during action learning. Meanwhile, continued training of the world expert on the DISR targets of robot clips absorbs the residual viewpoint gap caused by camera height and fisheye distortion.

\section{Experiments}
\label{sec:exp}

\subsection{Experimental Setup}
\label{sec:exp:setup}

\paragraph{EgoLynx.}
The quadruped baselines in Table~\ref{tab:main} are trained on the training split of EgoLynx-Robot, and only \method{} is additionally pretrained on EgoLynx-Web (Sec.~\ref{sec:model:training}). All methods are evaluated on the held-out test split of EgoLynx-Robot, which is sampled from the same teleoperation collection but shares no clips with the training split. We report open-loop displacement errors over the 1~s action chunk. We also report the proposed Quadruped Navigation Score (QNS), a pseudo-closed-loop metric inspired by NAVSIM \citep{navsim} that aggregates sub-scores covering safety, instruction compliance, comfort, and progress. Its computation is detailed in Appendix~\ref{app:qns}.

\paragraph{Perception comparison.}
Section~\ref{sec:exp:perception} compares DISR and task-specific OCC+BEV as future-prediction targets and policy inputs on EgoLynx-Robot, reporting QNS and end-to-end latency. OCC+BEV additionally requires Box/Line annotations and calibrated sensor point clouds for target construction. Point clouds are used offline in the target variant and supplied online in the OCC+BEV input variant. DISR is constructed from camera images. Latency includes online representation extraction, when required, and trajectory generation.

\textbf{NAVSIM.}
We follow the standard protocol of NAVSIM v1 \citep{navsim}, training on navtrain and evaluating on navtest with camera inputs only, and report the PDM score (PDMS).

\textbf{Implementation details.}
The understanding expert of \method{} is initialized from Qwen3-VL-2B \citep{qwen3vl}. Stage~1 and Stage~2 are trained for 20k and 5k steps, respectively, on 32 NVIDIA H800 GPUs. The driving model shares the same architecture as the quadruped model and differs only in $V$, $N_a$, and $\delta$.

\begin{table}[t]
    \centering
    \caption{Quantitative navigation results on the EgoLynx-Robot test split. ADE/FDE and AHE/FHE denote average/final displacement and heading errors, respectively. The pseudo-closed-loop metrics comprise no collision (NC), walkable-area compliance (WAC), time-to-collision (TTC), clearance (CL), ego progress (EP), instruction compliance (IC), and the Quadruped Navigation Score (QNS).}
    \label{tab:main}
    \renewcommand{\arraystretch}{1.0}
    \small
    \setlength{\tabcolsep}{2.8pt}
    \resizebox{\textwidth}{!}{
    \begin{tabular}{lccccccccccc}
        \toprule
        & \multicolumn{4}{c}{\textit{Open-loop}} & \multicolumn{7}{c}{\textit{Pseudo-closed-loop}} \\
        \cmidrule(lr){2-5}\cmidrule(lr){6-12}
        \textbf{Model}
        & \textbf{ADE$\downarrow$} & \textbf{FDE$\downarrow$}
        & \textbf{AHE$\downarrow$} & \textbf{FHE$\downarrow$}
        & \textbf{NC$\uparrow$} & \textbf{WAC$\uparrow$} & \textbf{TTC$\uparrow$} & \textbf{CL$\uparrow$}
        & \textbf{EP$\uparrow$} & \textbf{IC$\uparrow$}
        & \textbf{QNS$\uparrow$} \\
        \midrule
        \multicolumn{12}{l}{\textit{VLN-based policies}} \\
        InternNav~\citep{internnav}
        & 0.079 & 0.098 & 0.038 & 0.062
        & 0.945 & 0.955 & 0.871 & 0.815 & 0.893 & 0.818
        & 0.764 \\
        NaVILA~\citep{navila}
        & \underline{0.030} & \underline{0.059} & \underline{0.028} & \underline{0.048}
        & \underline{0.968} & \underline{0.975} & \underline{0.912} & 0.838 & \underline{0.941} & \underline{0.893}
        & \underline{0.843} \\
        OmniNav~\citep{omninav}
        & 0.090 & 0.152 & 0.033 & 0.054
        & 0.939 & 0.949 & 0.863 & 0.811 & 0.866 & 0.803
        & 0.743 \\
        StreamVLN~\citep{streamvln}
        & 0.036 & 0.067 & 0.032 & 0.053
        & 0.964 & 0.972 & 0.905 & \textbf{0.843} & 0.933 & 0.884
        & 0.833 \\
        \midrule
        \multicolumn{12}{l}{\textit{VLA-based policies}} \\
        LingBot-VLA~2.0~\citep{lingbotvla2}
        & 0.044 & 0.068 & 0.038 & 0.054
        & 0.958 & 0.966 & 0.893 & 0.829 & 0.926 & 0.863
        & 0.809 \\
        DepthVLA~\citep{depthvla}
        & 0.056 & 0.074 & 0.032 & 0.049
        & 0.952 & 0.961 & 0.882 & 0.825 & 0.915 & 0.842
        & 0.789 \\
        $\pi_{0.5}$~\citep{pi05}
        & 0.036 & 0.075 & 0.032 & 0.052
        & 0.961 & 0.969 & 0.899 & 0.834 & 0.922 & 0.874
        & 0.819 \\
        OpenVLA~\citep{openvla}
        & 0.041 & 0.068 & 0.030 & 0.051
        & 0.956 & 0.964 & 0.888 & 0.823 & 0.928 & 0.855
        & 0.802 \\
        \midrule
        \multicolumn{12}{l}{\textit{Ours}} \\
        \method{} w/o EgoLynx-Web
        & 0.034 & 0.062 & 0.030 & 0.052
        & 0.960 & 0.970 & 0.905 & 0.832 & 0.924 & 0.878
        & 0.823 \\
        \rowcolor{gray!12} \method{}
        & \textbf{0.027} & \textbf{0.052} & \textbf{0.024} & \textbf{0.041}
        & \textbf{0.982} & \textbf{0.979} & \textbf{0.928} & \underline{0.842} & \textbf{0.945} & \textbf{0.921}
        & \textbf{0.884} \\
        \bottomrule
    \end{tabular}
    }
\end{table}

\subsection{Main Results on Quadruped Navigation}
\label{sec:exp:main}

We compare \method{} with four VLN-based and four VLA-based policies on EgoLynx-Robot, all adapted to the same inputs and action chunk (Table~\ref{tab:main}). Among these policies, \method{} achieves the highest QNS, increasing it from NaVILA's 0.843 to 0.884, an absolute gain of 0.041 and a relative improvement of approximately 4.9\%. It also achieves the lowest reported ADE, FDE, AHE, and FHE. Relative to NaVILA, IC, TTC, and NC increase by 0.028, 0.016, and 0.014, respectively. Progress is slightly higher (0.945 vs.\ 0.941), while clearance remains close to the best baseline (0.842 vs.\ 0.843). The gains thus span instruction following and collision-related scores alongside maintained progress.

To assess the contribution of EgoLynx-Web pretraining within \method{}, we compare the full model with the variant trained on EgoLynx-Robot alone. Web pretraining raises QNS from 0.823 to 0.884, with improvements across all reported metrics, including instruction compliance and collision-related scores. Under the evaluated training protocol, these results support the contribution of action-free human-video pretraining to downstream navigation. The effects of individual pretraining signals are examined in Sec.~\ref{sec:exp:ablation}.

\begin{figure}[tb]
\centering
\includegraphics[width=0.98\textwidth]{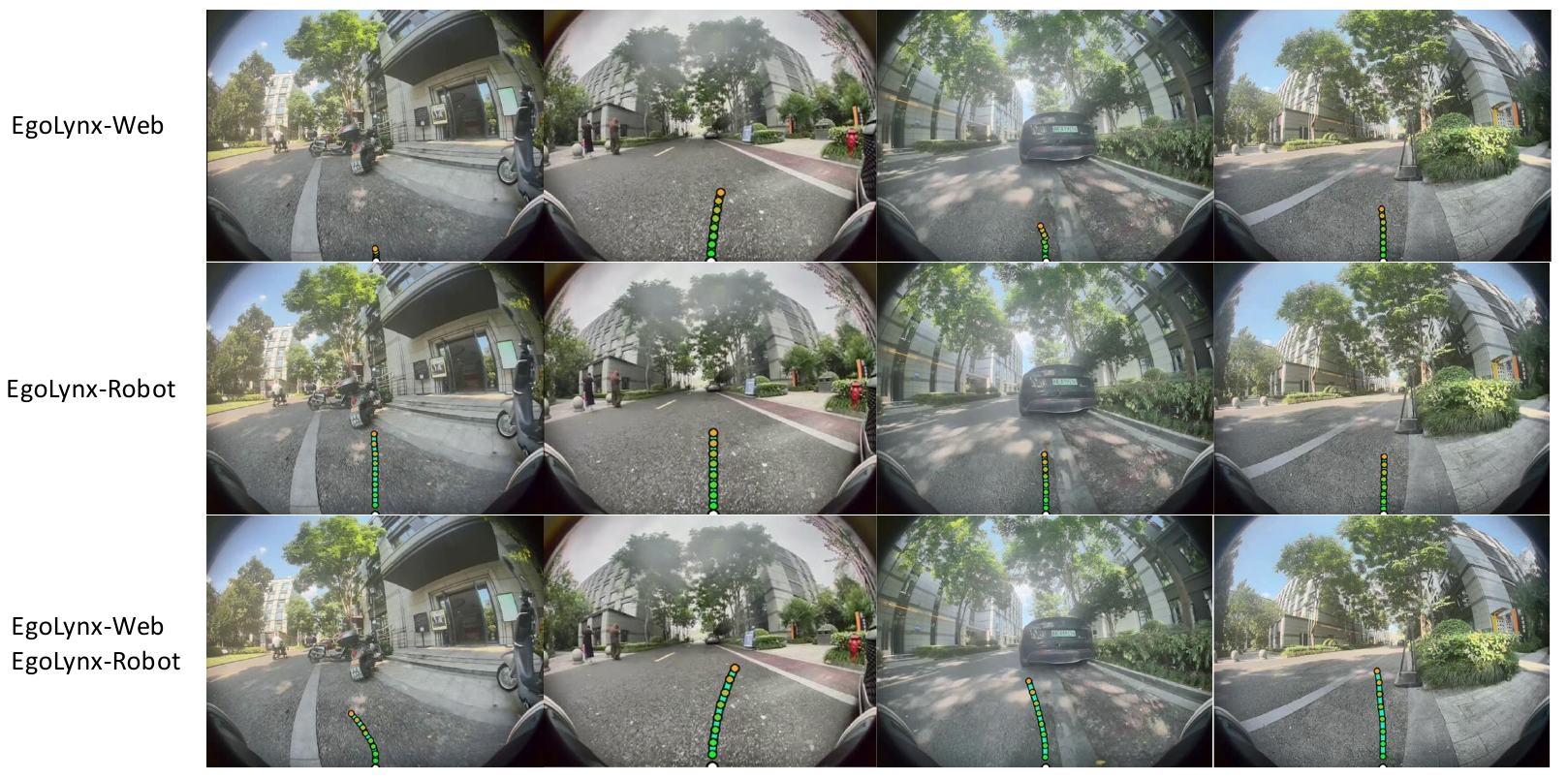}
\caption{\textbf{Predicted trajectories in unseen scenes.} Predictions are shown on replayed observations from four scenes (columns). The rows correspond to variants trained on EgoLynx-Web only (using DPVO pseudo-trajectories), EgoLynx-Robot only, or both.}
\label{fig:web_robot_qualitativ}
\end{figure}

Figure~\ref{fig:web_robot_qualitativ} illustrates web-only, robot-only, and jointly trained variants on replayed observations from unseen scenes. For the web-only variant, we obtain trajectory supervision from DPVO pseudo-trajectories, following \citet{citywalker}, without robot demonstrations. In these examples, its predictions steer around obstacles but are poorly calibrated to the quadruped's metric scale. The robot-only variant produces more suitably scaled trajectories but differs in its route choices around obstacles. In the third column, the jointly trained variant predicts a path that avoids an obstacle and then returns to the sidewalk. These examples illustrate differences in predicted paths within the replay setting.

\begin{table}[t]
\caption{\textbf{Ablation studies.} (a) Replacing DISR with alternative targets: Cosmos pixel latents \citep{cosmos}, DINOv3 \citep{dinov3} and V-JEPA~2.1 \citep{vjepa21} features, depth-augmented DINO video queries \citep{lingbotvla2}, and a FLARE-style embedding \citep{flare}. (b) Ablating DISR components. Architecture and two-stage training are fixed.}
\label{tab:world}
\centering
\small
\setlength{\tabcolsep}{3pt}
\begin{minipage}[t]{\textwidth}
\centering
\textit{(a) Alternative targets}\\[2pt]
\resizebox{\ifdim\width>\linewidth\linewidth\else\width\fi}{!}{%
\begin{tabular}{@{}lcccccc@{}}
\toprule
Target & ADE$\downarrow$ & FDE$\downarrow$ & NC$\uparrow$ & TTC$\uparrow$ & IC$\uparrow$ & QNS$\uparrow$ \\
\midrule
No world expert & 0.031 & 0.058 & 0.966 & 0.906 & 0.896 & 0.841 \\
Cosmos latent & 0.031 & 0.057 & 0.967 & 0.909 & 0.897 & 0.845 \\
DINOv3 & 0.030 & 0.056 & 0.971 & 0.914 & 0.903 & 0.856 \\
V-JEPA 2.1 & 0.029 & 0.055 & 0.973 & 0.916 & 0.906 & 0.861 \\
Depth + DINO video & 0.029 & 0.055 & 0.975 & 0.919 & 0.908 & 0.866 \\
FLARE-style & 0.030 & 0.057 & 0.970 & 0.912 & 0.905 & 0.853 \\
\rowcolor{gray!12} DISR (ours) & \textbf{0.027} & \textbf{0.052} & \textbf{0.982} & \textbf{0.928} & \textbf{0.921} & \textbf{0.884} \\
\bottomrule
\end{tabular}}
\end{minipage}\par\vspace{1em}
\begin{minipage}[t]{\textwidth}
\centering
\textit{(b) DISR components}\\[2pt]
\resizebox{\ifdim\width>\linewidth\linewidth\else\width\fi}{!}{%
\begin{tabular}{@{}lcccccc@{}}
\toprule
Variant & ADE$\downarrow$ & FDE$\downarrow$ & NC$\uparrow$ & TTC$\uparrow$ & IC$\uparrow$ & QNS$\uparrow$ \\
\midrule
No world expert & 0.031 & 0.058 & 0.966 & 0.906 & 0.896 & 0.841 \\
Segmentation only & 0.029 & 0.055 & 0.974 & 0.917 & 0.911 & 0.864 \\
Depth only & 0.029 & 0.054 & 0.976 & 0.921 & 0.909 & 0.868 \\
w/o relation layers & \underline{0.028} & \underline{0.053} & \underline{0.978} & \underline{0.922} & \underline{0.914} & \underline{0.873} \\
\rowcolor{gray!12} DISR (full) & \textbf{0.027} & \textbf{0.052} & \textbf{0.982} & \textbf{0.928} & \textbf{0.921} & \textbf{0.884} \\
\bottomrule
\end{tabular}}
\end{minipage}
\end{table}

\subsection{Ablation Study}
\label{sec:exp:ablation}
We evaluate prediction targets and model components on the EgoLynx-Robot test split. Additional results on expert layout, pretraining signals, and data scale are reported in Appendix~\ref{app:ablation}.

Removing the world expert lowers QNS to 0.841. With the expert retained, we compare DISR with five alternative targets under the same training data and schedule (Table~\ref{tab:world}(a)). Cosmos latents yield 0.845 QNS, close to the no-expert baseline. DISR achieves 0.884, compared with 0.866 for the strongest alternative, depth-augmented DINO queries, and improves IC and TTC by 0.013 and 0.009, respectively. These comparisons favor DISR among the evaluated generic visual targets and show that target choice affects downstream navigation performance.

The component ablations suggest complementary contributions from semantic and geometric cues (Table~\ref{tab:world}(b)). Depth alone yields higher NC and TTC, while segmentation alone yields slightly higher IC. Combining both cues with relational encoding improves all reported metrics over either individual cue. Removing the relation layers lowers QNS from 0.884 to 0.873 and reduces IC and TTC by 0.007 and 0.006, respectively. These results favor relational encoding within the evaluated architecture.

Stage~1 supervision also affects downstream performance (Table~\ref{tab:data}). Captions alone and DISR alone yield 0.843 and 0.866 QNS, respectively, while their combination reaches 0.884. DISR provides the larger individual gain, and captions add a further 0.018 QNS when combined with DISR, supporting complementary pretraining signals.

\begingroup
\input{sections/perception_comparison}

\endgroup

\subsection{Generalization to Autonomous Driving}
\label{sec:exp:navsim}

\begin{table}[htbp]
\caption{\textbf{Results on NAVSIM v1 navtest \citep{navsim}.} Sub-scores and PDMS of prior work are taken from the cited papers. ``C'' cameras, ``L'' LiDAR.}
\label{tab:navsim}
\centering
\small\renewcommand{\arraystretch}{0.9}
\setlength{\tabcolsep}{4pt}
\resizebox{\ifdim\width>\linewidth\linewidth\else\width\fi}{!}{%
\begin{tabular}{@{}lllcccccc@{}}
\toprule
Method & Input & Backbone & NC & DAC & TTC & Comf. & EP & PDMS \\
\midrule
TransFuser~\citep{transfuser} & C+L & ResNet-34 & 97.7 & 92.8 & 92.8 & 100 & 79.2 & 84.0 \\
DiffusionDrive~\citep{diffusiondrive} & C+L & ResNet-34 & 98.2 & 96.2 & 94.7 & 100 & 82.2 & 88.1 \\
WoTE~\citep{wote} & C+L & ResNet-34 & 98.5 & 96.8 & 94.9 & 99.9 & 81.9 & 88.3 \\
DiffusionDriveV2~\citep{diffusiondrivev2} & C+L & ResNet-34 & 98.3 & 97.9 & 94.8 & 99.9 & 87.5 & 91.2 \\
\midrule
PRIX~\citep{prix} & C & ResNet-34 & 98.1 & 96.3 & 94.1 & 100 & 82.3 & 87.8 \\
iPad~\citep{ipad} & C & ResNet-34 & 98.6 & \textbf{98.3} & 94.9 & 100 & \textbf{88.0} & 91.7 \\
\midrule
Fine-tuned Qwen2.5-VL~\citep{qwen25vl} & C & Qwen2.5-VL-7B & 97.8 & 92.1 & 92.8 & 100 & 78.3 & 83.3 \\
AdaThinkDrive~\citep{adathinkdrive} & C & InternVL3-8B & 98.4 & 97.8 & 95.2 & 100 & 84.4 & 90.3 \\
ReCogDrive~\citep{recogdrive} & C & InternVL3-2B + diffusion & 97.9 & 97.3 & 94.9 & 100 & 87.3 & 90.8 \\
\midrule
\method{} w/o world expert & C & Qwen3-VL-2B & 98.4 & 96.9 & 94.8 & 100 & 85.6 & 89.5 \\
\rowcolor{gray!12} \method{} (DISR target) & C & Qwen3-VL-2B & \textbf{99.0} & 98.0 & \textbf{96.9} & 100 & 87.8 & \textbf{92.1} \\
\bottomrule
\end{tabular}}
\end{table}

We further evaluate \method{} on NAVSIM to examine whether its design extends beyond quadruped navigation. Following proposal--selection planners \citep{hydramdp,wote}, the action expert decodes 64 candidate trajectories from learnable action queries, and a scorer supervised by the PDM sub-scores selects the final trajectory. All other components remain unchanged. \method{} achieves 92.1 PDMS, outperforming the compared planners, including LiDAR-based methods and VLM-based planners built on larger backbones (Table~\ref{tab:navsim}). Removing the world expert will lower PDMS to 89.5, with a notable drop in time-to-collision, which rewards anticipating the motion of other agents. The benefit of predicting future DISR therefore carries over to an embodiment whose viewpoint and dynamics differ from those of both humans and quadrupeds.

\section{Conclusion and Limitations}
\label{sec:conclusion}

\textbf{Conclusion.} 
This work studies how large-scale egocentric human videos can provide transferable knowledge for outdoor quadruped navigation. Our central thesis is that human-to-robot transfer should be grounded in scene structure rather than in pixels or actions. We instantiate this view with \method{}, a Mixture-of-Transformers model whose world expert learns to predict a domain-invariant scene representation of semantic, spatial, and relational structure from egocentric human videos, and curate EgoLynx, which pairs 10k hours of egocentric web videos with 2,000 teleoperated quadruped training clips. Ablations show that the prediction target matters: a reconstruction-oriented latent target provides little benefit over removing the world expert, whereas DISR performs best among the compared generic visual targets. Task-specific OCC+BEV targets yield close performance but require dedicated geometric labels and sensor point clouds. Online perception inputs improve QNS at increased latency, supporting DISR as a practical training target with camera-only policy inference. These results support structured future-scene prediction as a way to transfer useful knowledge from human videos to robot navigation.

\textbf{Limitations.} First, DISR is defined in the image plane, so the residual viewpoint gap still requires adaptation on robot data and precludes zero-shot cross-embodiment transfer. Second, DISR is supervised by pseudo-labels from off-the-shelf vision foundation models over a fixed category vocabulary, whose noise and coverage constrain the representation of long-tail scenes. Third, our quadruped evaluation relies on open-loop and pseudo-closed-loop metrics complemented by real-robot replay, leaving large-scale closed-loop deployment to future study. Future work will explore 3D semantic occupancy targets that preserve DISR's low annotation requirements, open-vocabulary prediction, and larger web corpora. More broadly, we envision world models that are pretrained on human videos and operate on scene structure as a shared interface for navigation across wheeled-legged robots and vehicles.

\begingroup
\small
\setlength{\bibsep}{2.5pt}
\bibliographystyle{plainnat}
\bibliography{references}
\endgroup

\clearpage
\appendix

\section{Additional Ablation Results}
\label{app:ablation}

This appendix reports the ablations of the expert layout, the Stage~1 pretraining signals, and the data scale summarized in Sec.~\ref{sec:exp:ablation}, evaluated on the EgoLynx-Robot test split under the same protocol.

\begin{table}[ht]
\begin{minipage}[t]{\textwidth}
\caption{\textbf{Ablation on the expert layout on the EgoLynx-Robot test split.} Metrics as in Table~\ref{tab:main}.}
\label{tab:layout}
\centering
\small\renewcommand{\arraystretch}{0.9}
\setlength{\tabcolsep}{2.5pt}
\resizebox{\ifdim\width>\linewidth\linewidth\else\width\fi}{!}{\begin{tabular}{@{}lcccccc@{}}
\toprule
Expert layout & ADE$\downarrow$ & FDE$\downarrow$ & NC$\uparrow$ & TTC$\uparrow$ & IC$\uparrow$ & QNS$\uparrow$ \\
\midrule
Full model & \textbf{0.027} & \textbf{0.052} & \textbf{0.982} & \textbf{0.928} & \textbf{0.921} & \textbf{0.884} \\
Dense backbone & 0.029 & 0.055 & 0.974 & 0.918 & 0.910 & 0.864 \\
Detached world loss & 0.029 & 0.054 & 0.977 & 0.921 & 0.913 & 0.870 \\
\bottomrule
\end{tabular}}
\end{minipage}\par\vspace{1em}
\begin{minipage}[t]{\textwidth}
\caption{Ablation on pretraining signals and data scale with Stage~2 fixed. The first block varies what EgoLynx-Web supervises. The last two blocks vary the amount of web and robot data. Metrics as in Table~\ref{tab:main}.}
\label{tab:data}
\centering
\small\renewcommand{\arraystretch}{0.9}
\setlength{\tabcolsep}{2.5pt}
\resizebox{\ifdim\width>\linewidth\linewidth\else\width\fi}{!}{\begin{tabular}{@{}lccccc@{}}
\toprule
Pretraining & Web (h) & Robot & ADE$\downarrow$ & FDE$\downarrow$ & QNS$\uparrow$ \\
\midrule
None & 0 & 2,000 & 0.034 & 0.062 & 0.823 \\
Captions only & 10k & 2,000 & 0.031 & 0.058 & 0.843 \\
DISR only & 10k & 2,000 & \underline{0.029} & \underline{0.055} & \underline{0.866} \\
Captions + DISR & 10k & 2,000 & \textbf{0.027} & \textbf{0.052} & \textbf{0.884} \\
\midrule
Captions + DISR & 1k & 2,000 & 0.031 & 0.058 & 0.845 \\
Captions + DISR & 3k & 2,000 & 0.030 & 0.056 & 0.862 \\
\midrule
Captions + DISR & 10k & 500 & 0.036 & 0.068 & 0.829 \\
Captions + DISR & 10k & 1,000 & 0.031 & 0.059 & 0.856 \\
\bottomrule
\end{tabular}}
\end{minipage}
\end{table}

\paragraph{Expert layout.}
We next examine two choices in the expert architecture (Table~\ref{tab:layout}): separating expert parameters and allowing the world objective to update the understanding expert. 
Replacing the Mixture-of-Transformers with a parameter-matched dense backbone reduces QNS by 2 points. 
Blocking gradients from the world loss to the understanding expert reduces QNS by 1.4 points. 
Both comparisons favor the full layout, supporting the use of separate experts and world-guided training of the understanding representation.

\paragraph{Pretraining signals.}
We examine which Stage~1 signals from EgoLynx-Web contribute to downstream navigation (Table~\ref{tab:data}). 
Relative to no pretraining, captions alone improve QNS by 2 points, DISR alone by 4.3 points, and their combination by 6.1 points. DISR provides the larger individual gain, while captions add a further 0.018 QNS when combined with DISR.

\paragraph{Data scaling.}
Finally, we vary the amounts of web video and robot data. 
QNS increases consistently over the evaluated range of web-data scales, from 0.845 at 1k hours to 0.862 at 3k hours and 0.884 at 10k hours.
With EgoLynx-Web pretraining, 500 robot clips yield slightly higher QNS than 2,000 clips without web pretraining (0.829 vs.\ 0.823). However, ADE and FDE remain higher (0.036/0.068 vs.\ 0.034/0.062). This comparison supports improved robot-data efficiency in terms of QNS while retaining a gap in open-loop trajectory accuracy.

\clearpage
\section{Quadruped Navigation Metric}
\label{app:qns}
\textbf{Trajectory imitation metrics.}
Given predicted and reference trajectories, ADE and FDE measure the mean
and final Euclidean position errors, respectively.
AHE and FHE measure the corresponding mean and final absolute heading
errors, with angular differences wrapped to $[-\pi,\pi)$.

\textbf{Pseudo-closed-loop replay.}

Annotated agents are replayed against the predicted ego trajectory and six interpretable sub-scores
measure No Collision (NC), Walkable-Area Compliance (WAC), Time-to-Collision (TTC), Clearance (CL), Ego Progress (EP), and Instruction Compliance (IC). They are combined as
\begin{equation}
    \mathrm{QNS}=\mathrm{NC}\times\mathrm{WAC}\times
    \frac{w_{\mathrm{EP}}\mathrm{EP}+w_{\mathrm{TTC}}\mathrm{TTC}+w_{\mathrm{CL}}\mathrm{CL}+w_{\mathrm{IC}}\mathrm{IC}}
    {w_{\mathrm{EP}}+w_{\mathrm{TTC}}+w_{\mathrm{CL}}+w_{\mathrm{IC}}}.
    \label{eq:qns}
\end{equation}
The replay is non-reactive.

\begin{itemize}

\item \textbf{No Collision (NC):}
NC is 0 for an at-fault collision with a person, rider, vehicle, animal, or
robot, 0.5 for a collision with an inanimate object, and 1 otherwise.
A collision is not at fault when the ego is stationary and another
object moves into it.

\item \textbf{Walkable-Area Compliance (WAC):}
WAC is 0 if the ego footprint crosses a curb or a road--sidewalk dividing line
outside a crosswalk and the recorded trajectory does not cross the same
boundary within the horizon. It is 1 otherwise.

\item \textbf{Time-to-Collision (TTC):}
TTC is 0 if 1-s constant-velocity propagation from any predicted step intersects
a replayed agent. It is 1 otherwise.

\item \textbf{Clearance (CL):}
$\min(1,d_{\min}/0.5\,\mathrm{m})$, where $d_{\min}$ is the minimum
distance to any person, rider, or animal.

\item \textbf{Ego Progress (EP):}
EP is the predicted progress projected onto the recorded path, normalized by the
recorded-path length and clipped to $[0,1]$. It is set to 1 when the recorded
motion is below 0.2\,m.

\item \textbf{Instruction Compliance (IC):}
IC checks instruction-specific trajectory geometry. Branches, U-turns, and
in-place rotation require at least $20^\circ$ of signed heading change.
Lateral or lane moves require signed dominant lateral velocity.
Reversing requires negative longitudinal velocity, while stationary or waiting
instructions require displacement below 0.2\,m. Obstacle avoidance requires
passage on the instructed side. Path following or continuing straight requires
continued motion when the operator moved.

\end{itemize}

\end{document}

%% file: sections/perception_comparison.tex
\subsection{Comparison with Task-Specific 3D Perception}
\label{sec:exp:perception}

We compare DISR with task-specific OCC+BEV perception on the held-out EgoLynx-Robot test split (Table~\ref{tab:perception-comparison}). Here, BEV comprises 3D object boxes and navigation-relevant road polylines predicted by a SparseDriveV2-like model \citep{sparsedrivev2} trained with Box/Line annotations. OCC is constructed by associating SAM semantic masks from past and current frames with calibrated sensor point clouds and voxelizing the semantically labeled points. We evaluate both representations as future-prediction targets and policy inputs.

\paragraph{As prediction targets.}
OCC+BEV and DISR yield close QNS values as future targets for the world expert (0.885 vs.\ 0.884), with reported latencies of 127 and 120~ms, respectively. Both variants construct targets offline and retain camera-only policy inference without an online target-construction front end. DISR achieves this navigation quality using segmentation and relative-depth foundation models, avoiding the task-specific Box/Line annotations and point clouds required to construct OCC+BEV targets.

\paragraph{As policy inputs.}
Using DISR as a policy input raises the reported QNS from 0.884 to 0.897, while latency increases from 120 to 220~ms. OCC+BEV input achieves a further 0.009 QNS gain over DISR input, accompanied by real-time sensor point clouds and a latency increase from 220 to 347~ms. Both input variants require online representation extraction, and latency includes the full perception and policy pipeline. Together, these results highlight the performance--latency trade-off and support DISR as a broadly obtainable training target that balances navigation quality, supervision requirements, and inference cost under the evaluated conditions.

\begin{table}[htbp]
\centering
\small
\setlength{\tabcolsep}{3pt}
\caption{DISR and OCC+BEV as targets or inputs on EgoLynx-Robot. Sensors are inference-time inputs. PC denotes sensor point clouds, used only offline in OCC+BEV-target. Box/Line labels are additional geometric supervision. Front end denotes online representation extraction. Latency includes the full perception and policy pipeline. The DISR-target QNS is from Table~\ref{tab:main}.}
\label{tab:perception-comparison}
\begin{tabular}{@{}>{\raggedright\arraybackslash}p{0.15\linewidth}>{\raggedright\arraybackslash}p{0.075\linewidth}>{\centering\arraybackslash}p{0.14\linewidth}>{\centering\arraybackslash}p{0.13\linewidth}>{\centering\arraybackslash}p{0.155\linewidth}>{\centering\arraybackslash}p{0.07\linewidth}>{\centering\arraybackslash}p{0.115\linewidth}@{}}
\toprule
Representation & Use & Sensors & \shortstack{Box/Line\\labels} & Front end & QNS $\uparrow$ & \shortstack{Latency\\(ms) $\downarrow$} \\
\midrule
DISR & Target & Camera & No & None & 0.884 & 120 \\
OCC+BEV & Target & Camera & Yes & None & 0.885 & 127 \\
DISR & Input & Camera & No & DISR extraction & 0.897 & 220 \\
OCC+BEV & Input & Camera + PC & Yes & OCC+BEV & 0.906 & 347 \\
\bottomrule
\end{tabular}
\end{table}